\documentclass[letterpaper, 10 pt, conference]{ieeeconf}  

\IEEEoverridecommandlockouts                              

\usepackage{graphicx}
\usepackage{epsfig}

\usepackage{hyperref}

\title{\LARGE \bf
Towards vision-based dual arm robotic fruit harvesting
}

\author{Ege Gursoy$^{1,2}$ Benjamin Navarro$^{1}$ Akansel Cosgun$^{2,3}$ Dana Kulić$^{2}$ Andrea Cherubini$^{1}$
\thanks{$^{1}$LIRMM, Univ. Montpellier, CNRS, Montpellier, France
        }%
\thanks{$^{2}$Monash University, Australia
        }%
\thanks{$^{3}$Deakin University, Australia
        }%
}

\begin{document}

\renewcommand{\baselinestretch}{0.9}

\maketitle
\thispagestyle{empty}
\pagestyle{empty}

\begin{abstract}
Interest in agricultural robotics has increased considerably in recent years due to benefits such as improvement in productivity and labor reduction. However, current problems associated with unstructured environments make the development of robotic harvesters challenging. Most research in agricultural robotics focuses on single arm manipulation. Here, we propose a dual-arm approach. We present a dual-arm fruit harvesting robot equipped with a RGB-D camera, cutting and collecting tools. We exploit the cooperative task description to maximize the capabilities of the dual-arm robot. We designed a Hierarchical Quadratic Programming based control strategy to fulfill the set of hard constrains related to the robot and environment: robot joint limits, robot self-collisions, robot-fruit and robot-tree collisions. We combine deep learning and standard image processing algorithms to detect and track fruits as well as the tree trunk in the scene. We validate our perception methods on real-world RGB-D images and our control method on simulated experiments.
\end{abstract}
\renewcommand{\baselinestretch}{0.8}
\section{INTRODUCTION} \label{introduction}
As one of the most important processes in agriculture, manual harvesting remains challenging due to its physically demanding nature.
With the expected global labor shortage\cite{karan2021resilience}, robotic harvesting becomes an important research area to explore. Robotic harvesting can improve productivity by: reducing labor costs, increasing yield quality and enabling better control of environmental implications \cite{kootstra2021selective}. 
The key challenges in robotic harvesting are:
\begin{enumerate}
\item \textbf{Fruit localization:} Determining the location of each fruit wrt the robot.
\item \textbf{Scene understanding:} Localizing potential obstacles in the scene such as branches, tree trunk, leaves etc. 
\item \textbf{Motion control:} Reaching the fruit while avoiding collisions, to detach and collect the fruit.
\end{enumerate}

\begin{figure}[t]
	\centering {\centering\includegraphics[width=0.95\columnwidth]{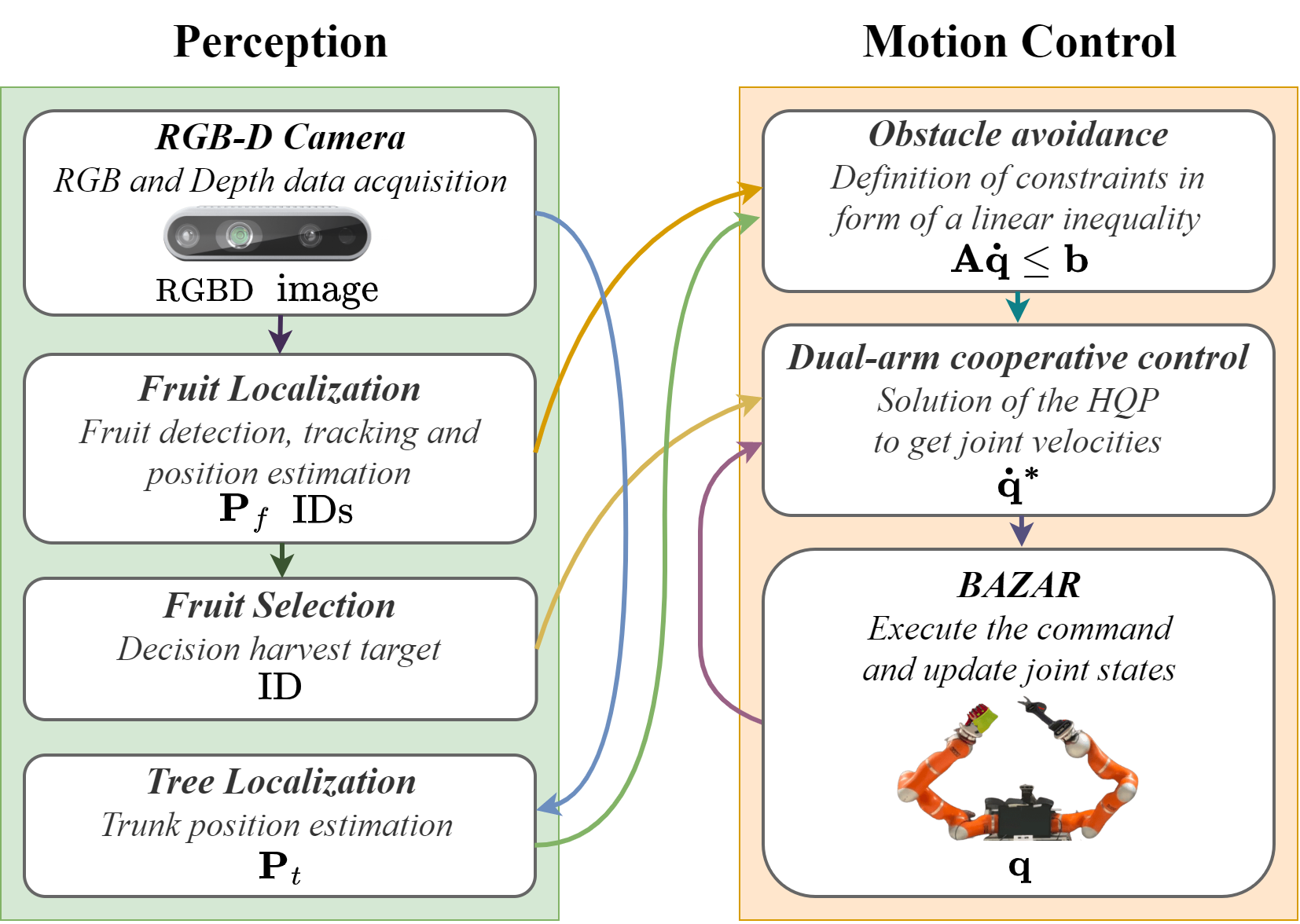}}
	\vspace{-0.1cm}
	\caption{Overview of our robotic harvesting system. Perception and Motion Control blocs are detailed in Section \ref{perception} and \ref{motion}, respectively.}
	\label{schema}
	\vspace{-0.6cm}
\end{figure}

In this work, we address  \textit{fruit localization} by detecting fruits in RGB-D images using machine learning and tracking them by assigning the same ID over successive frames. \textit{Scene understanding} is necessary to avoid robot collisions while reaching the fruit. 
On each image, we segment tree trunk and branches by combining a mask in Hue Saturation Value (HSV) color space with one in the depth image. We assume that tree branches are symmetric around their trunk. Then, we define a cylinder around the trunk, which the robot will avoid during operation. For \textit{motion control}, two arms (cutting and collecting) move towards the closest fruit, according to the cooperative task representation \cite{Chiacchio} while avoiding the trunk and other fruit. We are using a Hierarchical Quadratic Programming (HQP) based control algorithm to fulfill the constraints related to the robot and environment such as robot joint limits, robot self-collisions and robot-tree collisions. In the last phase, the cutting arm moves to the stalk bearing of the fruit (\textit{peduncle}) while the collecting arm moves below the fruit. The same cycle continues until there are no more reachable fruits left in the scene. An overview of our system is shown in Figure \ref{schema}

\begin{figure}[t]
	\centering {\centering\includegraphics[trim=0 2cm 0 0cm, clip,width=0.75\columnwidth]{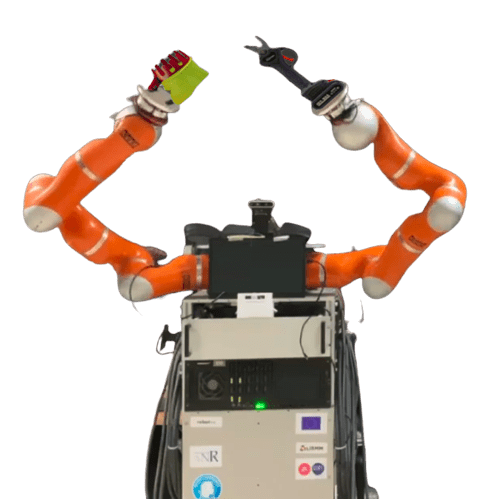}}
	\vspace{-0.1cm}
	\caption{Dual arm robot BAZAR \cite{cherubini2019collaborative} with cutting and collecting arms.}
	\label{bazar}
	\vspace{-0.5cm}
\end{figure}

A dual-arm robot compared to a single-arm, can perform more complex and delicate movements, ensuring careful handling of fruits, particularly those that are fragile or require precise detachment. The flexibility of the dual-arm increases the adaptability to different fruit types with the same set of end-effector tools. Moreover, the use of dual-arm robots enhances safety by minimizing damage to fruits and surrounding plant structures through increased precision.

The main contribution of this study is the preliminary validation of a dual-arm harvesting robot with vision and control, including : (i) real-time detection and tracking of fruit using deep learning methods (ii) dual-arm control in cooperative-task space based on Hierarchical Quadratic Programming (iii) collision avoidance of fruits and tree trunk by 2D RGB images and depth information (iv) generalization to round fruits grow on trees.

The remainder of the paper is structured as follows. A
review of the state of art in autonomous
fruit harvesting is given in Section \ref{work}. The perception of our robot is presented in Section \ref{perception}. The motion control and picking steps are explained in Section \ref{motion}. Finally, results of the experiments are presented in Section \ref{exp}, followed by a discussion and propositions for future work (Sec.~\ref{conclusion}).

\section{RELATED WORK} \label{work}
Robotic fruit harvesting has been demonstrated for various fruits such as apples \cite{davidson2017dual}, aubergines \cite{sepulveda2020robotic}, citrus \cite{mehta2014vision}, cucumbers \cite{van2002autonomous}, tomatoes \cite{ling2019dual}, strawberries \cite{hayashi2010evaluation} and sweet peppers \cite{lehnert2017autonomous}. Apart from \cite{davidson2017dual}\cite{sepulveda2020robotic}\cite{ling2019dual}, the cited systems only use one robotic arm, whose end-effector is responsible for both grasping and detaching the fruit, using rigidly linked gripper and cutter. This rigid link is specific for each fruit, limiting versatility. To address the broad diversity of fruit shapes and sizes, we propose the use of two arms.


Detecting fruits and locating them is the first challenge for the robot. 
Deep learning detectors are popular among agricultural robotics researchers. In \cite{santos2020grape}, Santos et al. introduce an interactive instance mask generation on RGB images based on graph matching and compare the performance of learning based object detection algorithms on grapes. Kang et al. \cite{kang2020fast} introduce an automatic label generation module for the detection of apples in RGB images of orchards. In \cite{koirala2019deep}, Koirala et al. investigate popular object detection frameworks for mango detection and design a custom detection network. Liu et al. \cite{liu2018visual} follow a multi-class approach in the detection of citrus by splitting their samples into classes. Kirk et al. \cite{kirk2020b} use a custom neural network  which takes a 6-channel RGB + CIELAB representation as input. In \cite{ganesh2019deep} Ganesh et al. train a neural network with RGB and HSV 6-channel input to detect oranges. However, the majority of these approaches have complex pipelines which are difficult to train and hard to generalize to other crop types.

Fruit tracking is another key requirement for
autonomous fruit harvesting. 
Roy et al. \cite{roy2016surveying} use apple contours in an Structure from Motion (SfM) pipeline, to generate dense matches and reconstruct the apples in 3D. However, SfM may have difficulty distinguishing between fruit and non-fruit objects that have similar contours. Das et al. \cite{das2015devices} use a Support Vector Machine to detect fruits, optical flow to associate the fruits between frames. However, optical flow may suffer in case of occlusions.  Liu et al. present a convolutional network to separate fruit and non-fruit pixels and use a Kalman Filter based Hungarian algorithm to correct Kanade–Lucas–Tomasi tracker which is computationally expensive.

Motion control is an equally important part of the harvesting robot. Ling et al. \cite{ling2019dual} propose a dual-arm robot for harvesting tomatoes. The vacuum cup-type tool attached to one of the end-effectors grasps the target fruit, while the other arm (equipped with a cutter) cuts the stem. However, the vacuum cup-type tool may not work well with fruits that have softer or more delicate skin, which may get damaged during the harvesting process. Sepulveda et al. \cite{sepulveda2020robotic} introduce a dual-arm aubergine harvesting robot configured similarly to humans. The planning algorithm decides the movement modes which are the simultaneous harvesting of two fruits or harvesting a single fruit with one arm, depending on the fruit location. 
Davidson et al. \cite{davidson2017dual} propose a dual-arm coordination strategy for fruit collection to reduce the harvesting cycle time. Rather than displacing the fruit to a bin, the picking arm drops the apple into a robotic collector near the point of fruit detachment. Their approach is hard to generalize to different fruits. Additionally, the dropping of the fruit may result in damage or bruising, which can reduce the quality and value of the harvested fruit.

\section{PERCEPTION} \label{perception}
Here, we present our perception system for detecting, tracking and localizing fruits, the fruit selection strategy, and the tree location estimation methodology.

\begin{figure}[t]
	\centering {\centering\includegraphics[width=0.85\columnwidth]{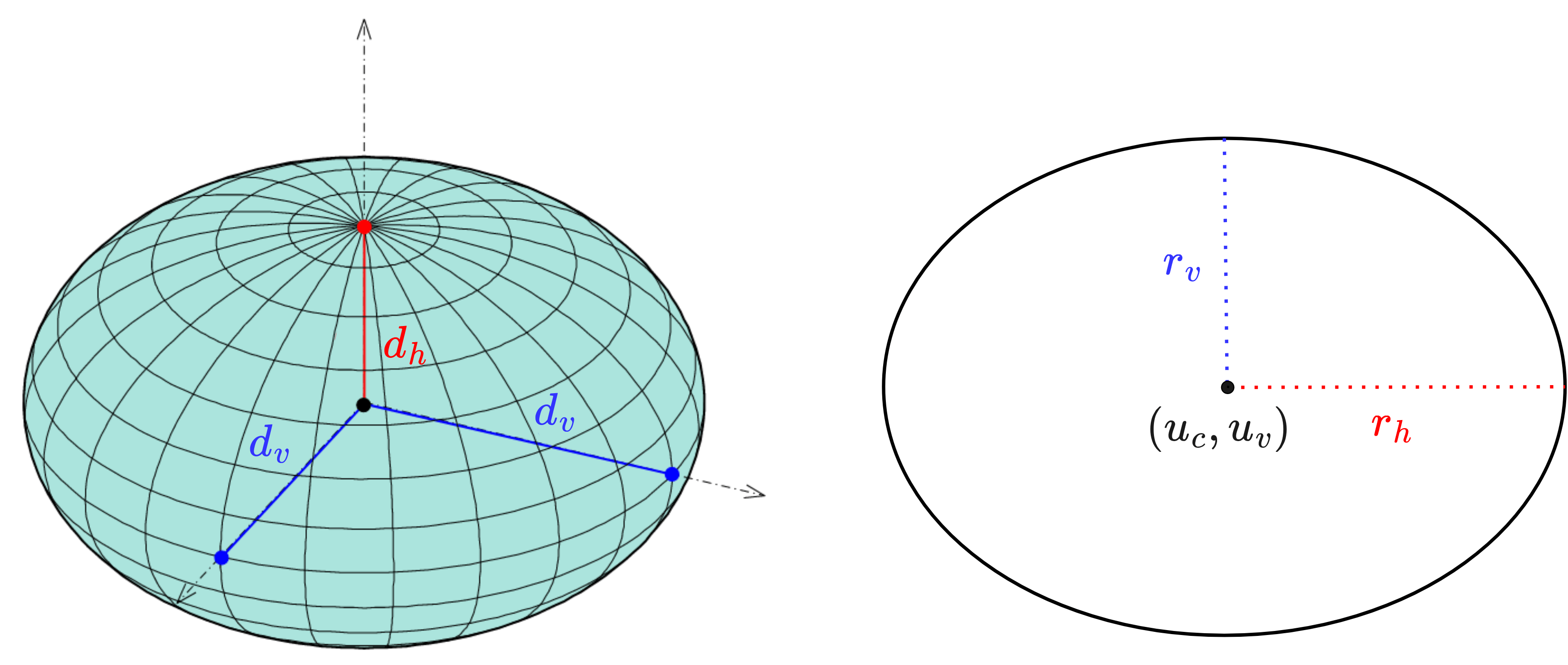}}
	\vspace{-0.1cm}
	\caption{Fruit ellipsoid assumption (left) and its ellipse projection (right).}
	\vspace{-0.2cm}
	\label{ellipsoid_ellipse}
\end{figure}

In this work, we assume that fruits are ellipsoids (see Figure \ref{ellipsoid_ellipse}), which appear in a 2D image as ellipses with axes aligned with the image height and width. The ellipsoid surrounding the fruit has a circular symmetry i.e. it has two equal semi-diameters along the $x$ and $z$ axes in the camera frame. Since most round fruits (e.g., apples, lemons) have this property, we can assume that our approach is valid in those cases.

\subsection{Fruit Localization} \label{localization}
\begin{figure}[t]

	\centering {\centering\includegraphics[width=0.85\columnwidth]{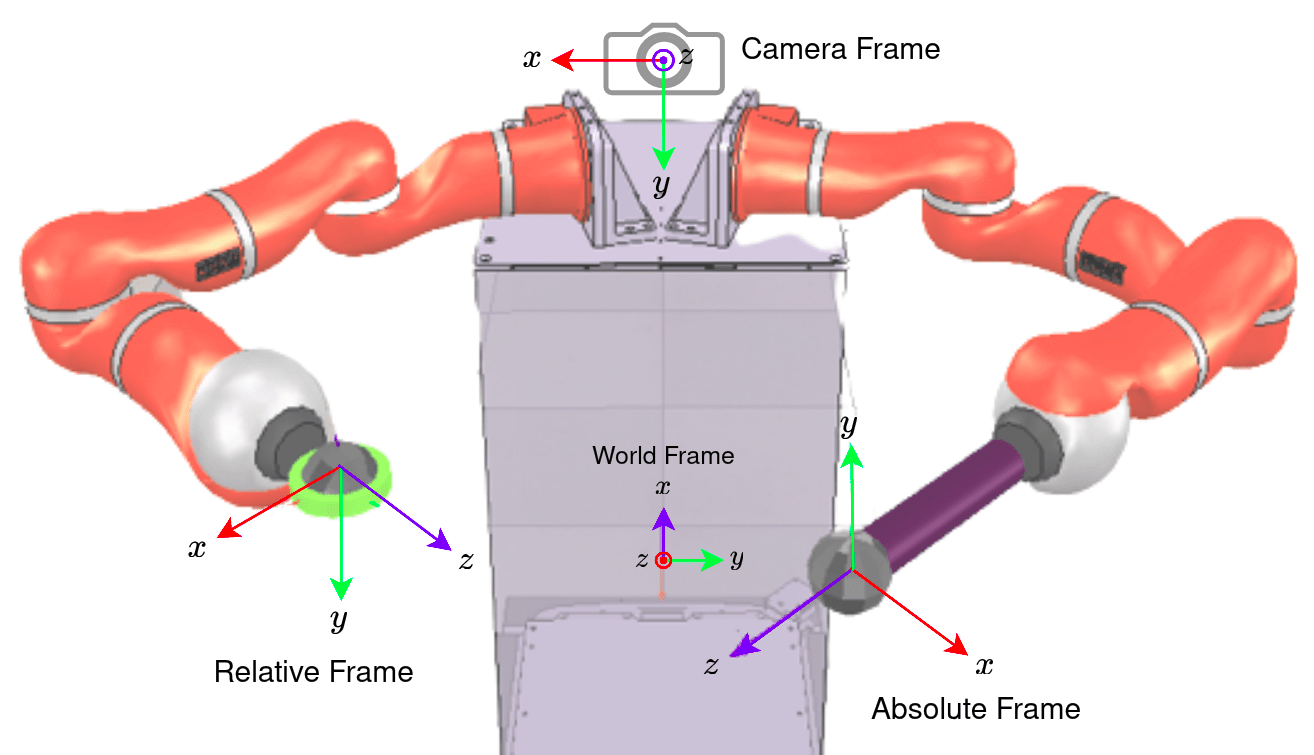}}
	\vspace{-0.1cm}
	\caption{Coordinate frames definitions. Absolute Frame links to fixed World Frame, while Relative Frame links to moving Absolute Frame.}
	\vspace{-0.6cm}
	\label{frames}
\end{figure}

Fruit detection consists in determining the pixel coordinates of the corners of the rectangle which encloses the fruits in the image. The specifications of each detected fruit are fed to the tracking algorithm, which assigns a unique ID to each fruit and correlates the same fruit data in different instances to determine whether or not new detected fruits can be associated with the existing IDs. Finally, fruit localization refers to calculating the 3D spatial position of the fruits the in camera frame shown in Figure \ref{frames}.

We use a deep learning object detection algorithm, YOLOv5 \cite{glenn_jocher_2022_7002879} because of its speed and accuracy. We conduct domain adaptation and transfer learning of YOLOv5. We initialize the network with pre-trained weights on COCO dataset \cite{lin2014microsoft} to benefit from their feature extractors and use 300 orange images from AgriPlant Dataset \cite{pawara2017comparing} to train. We split the dataset in 70:30 ratio for training and validation. The fruit tracking task relies on DeepSORT \cite{wojke2017simple}, an improved version of SORT \cite{bewley2016simple}, based on frame-to-frame association with a deep association metric. 


With the detected fruits in bounding boxes, fruit 3D positions are computed by incorporating the depth information from the RGB-D camera. We denote $(u_{TL}, v_{TL})$ and $(u_{BR}, v_{BR})$ the top-left and bottom-right pixel coordinates of the bounding box detected by YOLO. Fruit center in pixels is calculated as:
\begin{equation}
\left( u_C, v_C \right)= \left( \frac{u_{BR} – u_{TL}}{2}, \frac{v_{BR} – v_{TL}}{2}\right)
\end{equation}

Under this hypothesis, we compute the horizontal and vertical semi-axes of the ellipse surrounding the fruit $r_h = u_{BR} - u_C$ and $r_v = v_{BR} – v_C$

We denote the projection of a pixel in color information $(u,v)$ to a 3D position in camera frame $(x,y,z)$ as follows:
\vspace{-0.2cm}
\begin{equation} \label{eq_proj}
x,y,z = \mathrm{proj}(u,v) 
\vspace{-0.01cm}
\end{equation}

For each detected fruit, we compute its center $\mathbf{P}_f = x_f,y_f,z_f$ and distance to the peduncle from the center $d_v$. 
Under the ellipsoidal fruit assumption, the point visible at $(u_C, v_C)$ is the nearest to the camera optical center. We denote its depth $z_{min}$. the fruit depth is calculated by:
\vspace{-0.01cm}
\begin{equation} \label{eq_zf}
z_f = z_{min} + \mathrm{dist}[\:\mathrm{proj}(u_C, v_C),\; \mathrm{proj}(u_C + r_h, v_C)\:] 
\vspace{-0.01cm}
\end{equation}
with $\mathrm{dist}$ as the distance between two points in 3D Cartesian space.

The distance to the peduncle from the center is considered to be equal to the 3D transformation of the vertical semi-axis of the ellipse surrounding the fruit: 
\vspace{-0.1cm}
\begin{equation}
d_v = \mathrm{dist}[\:\mathrm{proj}(u_C, v_C),\; \mathrm{proj}(u_C, v_C + r_v)\:] 
\vspace{-0.01cm}
\end{equation}
These results will be used in the following to determine the cutting point.

\subsection{Fruit Selection} \label{selection}
Fruit selection describes the process of determining the best candidate fruit to harvest. We compute the distance of the fruits in the scene to the optical center of the camera using the color and depth information as described in Section \ref{localization}. For each image, these distances with their associated IDs are stored in a list. We update the list if the position error between the previous state of a fruit ID is smaller than a threshold, otherwise the previous state is kept. This ensures a robust approach to counter temporary ID errors caused by occlusions. We delete IDs if they are not detected in the scene for 100 image frames. Finally, we select the ID with smallest distance for harvesting.

\subsection{Tree Localization} \label{treestim}
We must estimate the tree position to avoid collisions between the robot and trunk to prevent damaging the plant or the robot while reaching for a fruit. 
Instead of machine learning, we use a simpler, computationally efficient method, based only on color and depth information. 

A HSV color space segmentation mask and a depth mask separate tree branches and trunk from the rest of the scene while assuming that they are not occluded in the scene. We convert the RGB information into the more consistent HSV space. The HSV space is chosen since its dependence upon hue is assigned to a single value (H) rather than across all three RGB values. 

On top of the HSV mask, we create a second mask using the depth information to filter out pixels that exceed a certain depth threshold. The secondary mask ensures pixels in the HSV range are not too far away and do not correspond to 3D points which do not belong to the tree.


The mean of the $(u,v)$ pixel coordinates in the RGB image and the depth value of the remaining pixels give the position of the tree. We denote the estimation of 3D coordinates of the tree in camera frame as $\mathbf{P}_t = x_{t},y_{t},z_{t}$. The mean depth of the remaining pixels after the mask gives us $z_{t}$, while $x_{t}$ and $y_{t}$ are calculated using (\ref{eq_proj}).

\section{MOTION CONTROL} \label{motion}
For a dual-arm fruit harvesting robot, both manipulator arms need to approach the fruits located at different positions with high accuracy and flexibility, while safely manipulating the object. To achieve this, we design a dual-arm control strategy paired with a cooperative representation method.

\subsection{Dual-arm cooperative control} \label{cooperative}
We describe our fruit picking task using the cooperative task representation \cite{adorno2010dual}, which fully characterizes the operational space for dual-arm control. This representation allows us to specify the operations in terms of an absolute task with pose attached to a fixed world frame and a relative task attached to the absolute task, as shown in Figure \ref{frames}.
The main reasons for using this representation are the ability to adapt control algorithms originally designed for single manipulators (e.g., quadratic programming based approaches) to dual-arm robots, and to perform safe dual-arm object manipulation without hitting the obstacles nor damaging the fruits. Our tasks include a cutting phase to detach the fruit from its stem and a collecting phase to catch the falling fruit. We believe that this representation is a proper fit to our tasks since the cutting is best to describe in a fixed frame while catching is more appropriately described in the relative frame.

We define fruit cutting as the absolute task which controls the motion of the robot in the workspace and fruit collecting as the relative task, which regulates the relative motion between the two arms.

Let us now formally define these absolute and relative tasks. We define a desired pose of the absolute task $\mathbf{x}_a = [x_a,y_a,z_a,\phi_a,\theta_a,\psi_a]$ for the cutting (left) arm in a fixed frame, (in our case, the camera frame) and a desired pose of the relative task $\mathbf{x}_r = [x_r,y_r,z_r,\phi_r,\theta_r,\psi_r]$ for the collecting (right) arm w.r.t. the frame attached to the cutting (left) arm during the harvesting phase.
Once we select the fruit to harvest in Section \ref{selection}, both arms should go towards that fruit. While harvesting, the tip of the cutting tool should be placed at the peduncle to detach the fruit while the center of the collecting tool is slightly below the fruit to catch it with the basket when it falls. We compute the peduncle location using the vertical semi-axis $d_v$ and fruit position $\mathbf{P}_f$ as described in Section \ref{localization}.  
\begin{figure}[t]
	\centering {\centering\includegraphics[trim=0 6cm 0 3cm, clip,width=0.9\columnwidth]{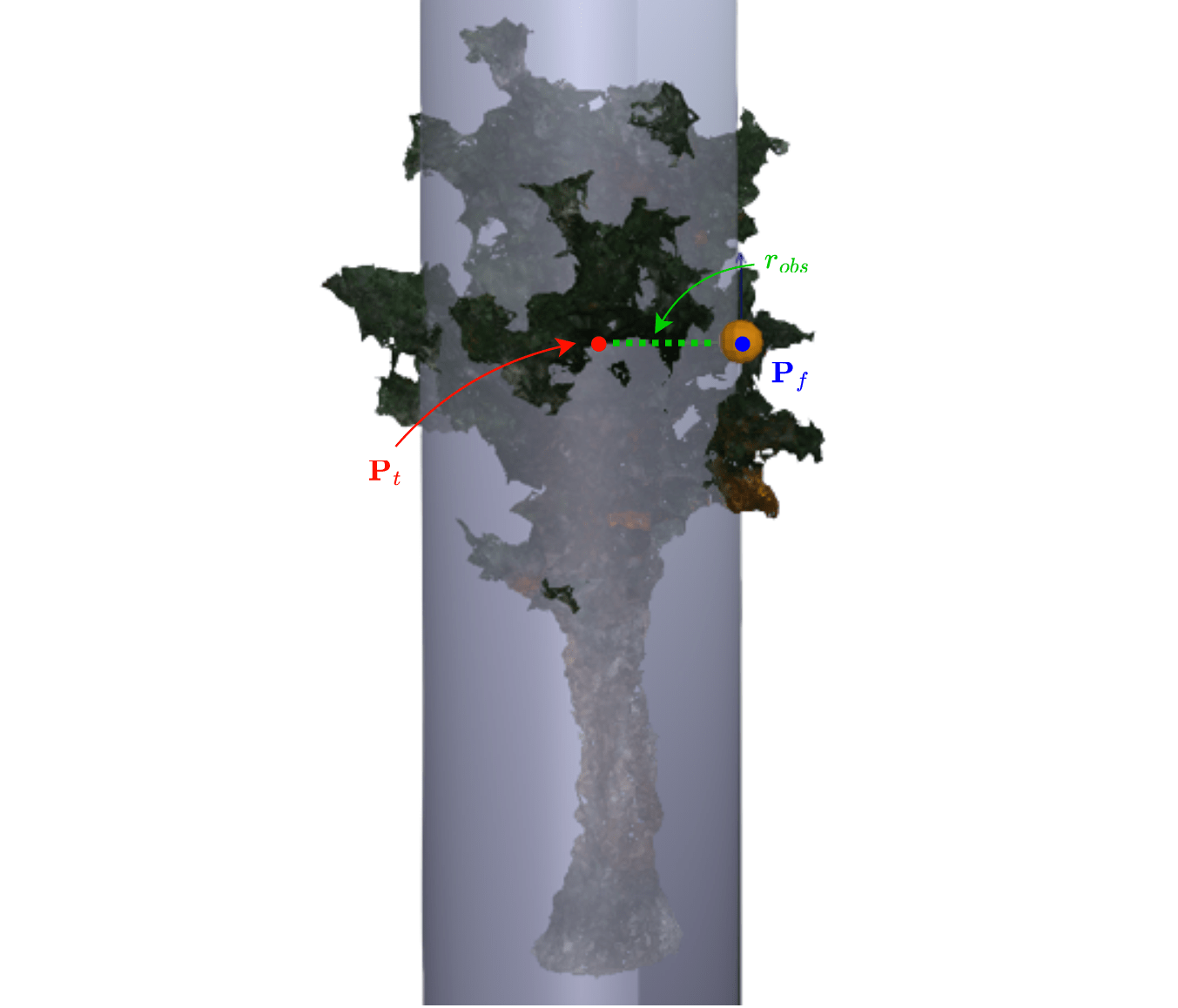}}
	\vspace{-0.1cm}
	\caption{Trunk obstacle cylinder in gray and fruit obstacle in orange.}
	\vspace{-0.4cm}
	\label{obstacl}
\end{figure}
As for the orientations, we do not impose any hard constraint on $\theta_a$ to maximize the reachable points in the workspace since cutting only requires the tip of the cutting tool to reach the peduncle. $\theta_r$ is constrained to $\pi/2$ radian, resulting in $\pi/2$ radian angle between two arms at all times, to prevent collisions between the arms. Other orientations are set to stay unchanged from the natural pose of the robot while its arms are in front.
The desired poses to reach a fruit relative to the camera frame are described by:
\vspace{-0.1cm}
\begin{IEEEeqnarray}{c} \label{des_abs_rel}
\mathbf{x}_a = [x_f ,y_f - d_v - d_s, z_f, 0, *,\pi] \nonumber\\
\mathbf{x}_r = [0, -d_v - d_s, 0, 0, \pi/2,\pi]
\vspace{-0.1cm}
\end{IEEEeqnarray}
with $d_s$ a security distance while cutting the fruit, and $*$ indicating unconstrained task component. 

In the cooperative dual-arm task representation, both the relative and absolute tasks have to be realized at the same time. However, a relevant choice is to prioritize the absolute task over the relative one. Indeed, it is crucial to fulfill the absolute task since the cutting arm position must be more precise than the collecting arm. 

Thus, we use Hierarchical Quadratic Programming (HQP) as described in \cite{tarbouriech2022admittance}. It consists in solving a sequence of quadratic programming problems for which the solution space of a given task is restricted to the null space of higher priority tasks. This ensures that low priority tasks to do not interfere with the others. The relationship between joint velocities $\mathbf{\dot{q}}$ and task space velocities $\mathbf{\dot{x}}$ is given by $\mathbf{\dot{x}} = \mathbf{J}\mathbf{\dot{q}}$ with $\mathbf{J}$ the Jacobian matrix.

The Inverse Kinematics module based on HQP outputs the joint velocity vector which will be sent to the robot. HQP includes a set of hard constraints to be fulfilled at all times, defined both at joint space level (joint limits) and task space level (collision avoidance). We define the set of constraints under the standard form of linear inequality equations:
\vspace{-0.1cm}
\begin{equation} \label{eq_ineq}
\mathbf{A}\mathbf{\dot{q}} 	\leq \mathbf{b}
\vspace{-0.1cm}
\end{equation}
where $\mathbf{A}$ is the linear coefficients matrix and $\mathbf{b}$ the
constant vector in the inequality constraints.

Therefore, in the cooperative dual-arm task space and taking into account constraints, the highest and lowest priority tasks are solved through the following optimization problems:
\vspace{-0.3cm}
\begin{IEEEeqnarray}{c}
\mathbf{\dot{q}^*_\mathit{a}} = \arg\min_\mathbf{\dot{q}}  ||\mathbf{J_\mathit{a}}\mathbf{\dot{q}}-\mathbf{\dot{x_\mathit{a}}}||_2 \nonumber\\
\textit{s.t.} ~~~~\mathbf{A}\mathbf{\dot{q}} 	\leq \mathbf{b}\nonumber\\
\mathbf{\dot{q}^*_\mathit{a,r}} = \arg\min_\mathbf{\dot{q}}  ||\mathbf{J_\mathit{r}}\mathbf{\dot{q}}-\mathbf{\dot{x_\mathit{r}}}||_2\\
\textit{s.t.}~~~~ \mathbf{A}\mathbf{\dot{q}} 	\leq \mathbf{b}\nonumber\\
~~~~~~~~~~~~\mathbf{J_\mathit{a}\dot{q} = J_\mathit{a}\dot{q}^*_\mathit{a}}\nonumber
\vspace{-0.1cm}
\end{IEEEeqnarray}
where the $\mathbf{a}, \mathbf{r}$ subscripts define the high priority (absolute) and low priority (relative) tasks and $\mathbf{\dot{q}}$ refers to the vector concatenating the joint velocities of the two arms.

\subsection{Obstacle Avoidance}\label{obstacle}
To prevent collisions with the fruit and branches which can damage the robot and the plant, we approach and leave the fruit from the front before and after the harvesting phase. We define additional tasks for before and after harvesting, named \textit{approach} and \textit{leave} tasks. The desired poses for the \textit{approach} and \textit{leaving} task are as (\ref{des_abs_rel}) except $z_a = z_f - d_z$ with $d_z$ as a parameter defining the distance to the fruit before passing to the harvesting phase.

To prevent any dangerous situation, we define the branches and trunk of the tree as an obstacle to avoid. In a real-world scenario, the branches of a tree have very complex shapes. Hence, we opt for a computationally efficient obstacle which covers the tree branches and trunk. We consider an infinite height cylinder collision object around the trunk of the tree with a variable radius to prevent the robot from entering the area between the selected fruit and the trunk. The cylinder radius depends on the fruit position. Variable radius provides us a safe approach distance towards the fruit which prevents the robot from entering the area behind the fruit. Let $r_{obs}$ be the radius of the cylinder. We place the center of the cylinder at $\mathbf{P}_{t}$. The height of the cylinder lays on the $y$ axis in the camera frame. The radius is defined as $r_{obs} = \mathrm{dist}[\:\mathbf{P}_{f} ,\;\mathbf{P}_{t}\:] - d_h$
with $\mathrm{dist}$ the distance between two points in 3D Cartesian space.
For each new fruit, $r_{obs}$ is 
updated. 

Since we model each fruit as an ellipsoid with circular section in the $xz$ plane, the fruit is characterized by: its position $\mathbf{P}_{f}$ and semi-axis lengths $d_h$ and $d_v$.

The tree and fruit obstacles are shown in Figure \ref{obstacl}. It is important to note that since the fruit and tree positions 
are calculated in real-time, the location and size of the obstacles is also updated accordingly.

After modeling the obstacles, a collision avoidance constraint is generated for all pairs of potentially colliding parts which are separated by a distance $d \leq d_i$. The obstacle avoidance strategy is expressed as linear constraints using Velocity Damper \cite{faverjon1987local}. The definition of the Velocity Damper is given by:
\vspace{-0.1cm}
\begin{equation} \label{damper}
\dot{d}^* = -\xi \frac{d-d_s}{d_i-d_s}, d \leq d_i
\vspace{-0.1cm}
\end{equation}

\begin{figure}[t]
	\centering {\centering\includegraphics[trim=0 0cm 0 1cm, clip,width=0.85\columnwidth]{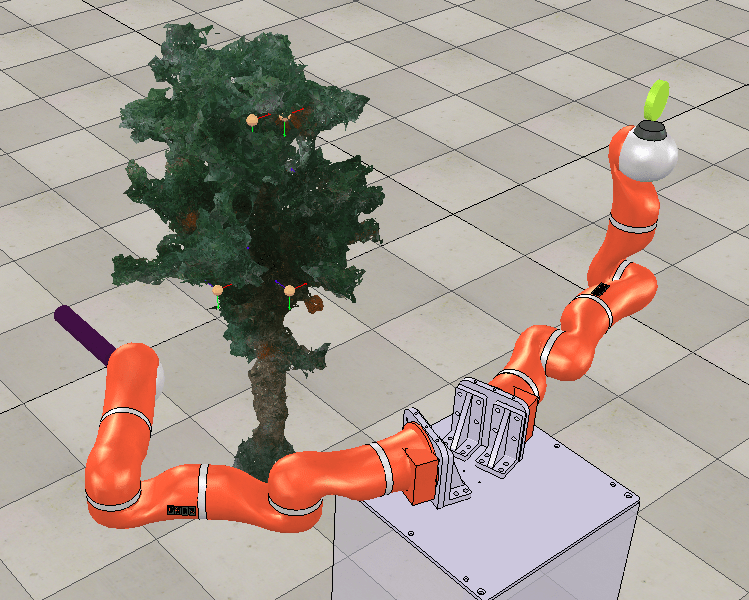}}
	\vspace{-0.1cm}
	\caption{Simulation with scanned tree, detected oranges and BAZAR.}
	\vspace{-0.6cm}
	\label{scene}
\end{figure}

with $d_i$ the activation distance, $d_s$ the safety distance, $d$ the current distance between bodies, and $\xi$ the convergence speed parameter.
The derived inequality constraint is given by $\dot{d}\geq\dot{d}^*$. When two bodies are separated by a distance smaller than $d_i$, their relative speed becomes bounded. The motion is slowed
down until $d_s$. If $d$ becomes larger than 0, the only possible motion between the bodies is to maintain or increase their relative distance. The relative distance can be expressed as:
\vspace{-0.1cm}
\begin{equation} \label{rel_dist}
\dot{d} = \mathbf{nJ_\mathit{p}\dot{q}_\mathit{d}}
\vspace{-0.1cm}
\end{equation}
with $\mathbf{n}$ the normalized vector joining the points on the objects and $\mathbf{J_\mathit{p}}$ the Jacobian matrix computed at one object and takes as reference
the other one.
Using (\ref{damper}) and (\ref{rel_dist}), we can specify
a constraint in function of joint velocities under the standard form of linear inequalities (\ref{eq_ineq}) with $\mathbf{A} = \mathbf{nJ_\mathit{p}}$ and $\mathbf{b} = \dot{d}^*$.
 
\subsection{Detaching and Collecting}
Although we do not cite any specific tools, they can be imagined as shears to detach the fruit and a basket to catch them while falling. Once the desired poses are reached by both arms, the cutting tool is triggered and the collecting arm waits the falling fruit. 
Then, we select a new fruit with the method explained in Section \ref{selection} and repeat the process.


\section{EXPERIMENTS} \label{exp}
We conduct a case study to validate our methodology on artificial orange trees in our laboratory. The fruit detection and tracking strategies are tested on images coming from the RGB-D camera, in real-time. We create a simulation scene with the 3D model of our orange tree as well as the dual arm robot kinematic model to validate our motion control strategy. 
The video of the experiments attached to the paper is also available online at: \url{https://youtu.be/ObCR5biqlOM}.


\begin{figure}[t]
	\centering {\centering\includegraphics[width=0.9\columnwidth]{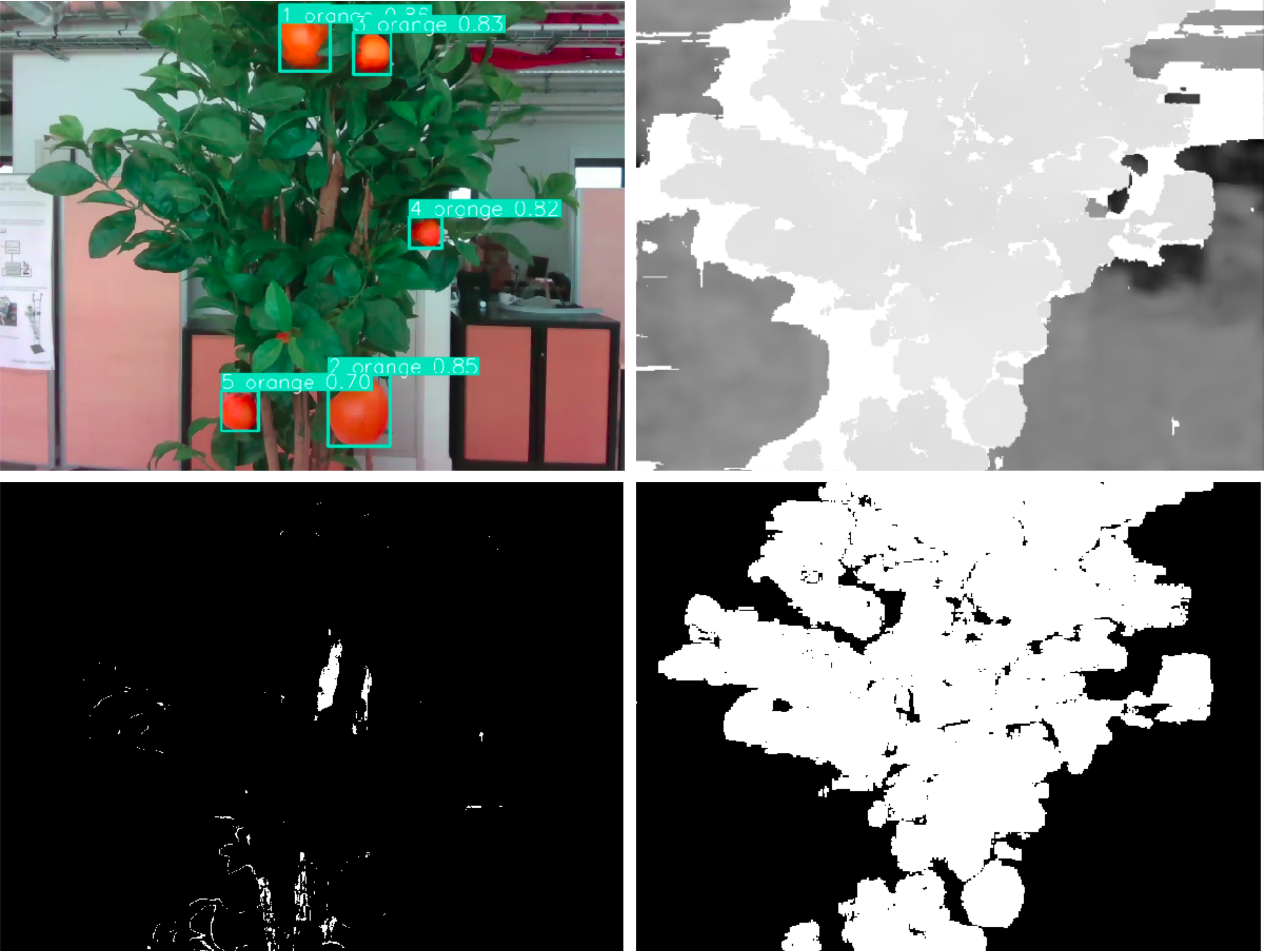}}
	\vspace{-0.1cm}
	\caption{RGB image with the fruit detection and depth information are shown in top-left and top-right. Bottom-left and bottom-right images show the HSV and depth masks.}
	\vspace{-0.6cm}
	\label{tree}
\end{figure}

\subsection{Platform Design}
The setup consists of dual-arm mobile cobot BAZAR \cite{cherubini2019collaborative} shown in Figure \ref{bazar} equipped with two 7-DOF Kuka LWR4 arms
and an Intel Realsense D435 RGB-D camera.
We implement the robot controller 
with the Robot Kinematic Control Library (RKCL). 
Hyper-parameters for YOLO are set to default, for DeepSORT \textit{model type}: OSnet x0.25 \cite{zhou2019omni}, \textit{cosine distance}: 0.2, \textit{overlap distance}: 0.5, \textit{age}: 300, \textit{n init}: 10, \textit{nn budget}: 30. Intel realsense SDK is used for the image and camera 
parameters acquisition 
and the 2D to 3D projection functions. We use OpenCV \footnote{https://opencv.org} for image processing operations. HSV values used for the segmentation mask are \textit{H}: 10-20, \textit{S}: 70-130, \textit{V}: 80-140 according to OpenCV's HSV conventions. 
Finally, CoppeliaSim\footnote{https://www.coppeliarobotics.com} is used to simulate the robot in real-time.

\subsection{Autonomous Harvesting}
Fruit detection and tracking are realized using the camera on BAZAR. Then, to simulate robot actions, we created a CoppeliaSim scene shown in Figure \ref{scene}. The scene includes a functional BAZAR robot and a 3D scan of the artificial orange tree. We detect the oranges in our lab using the RGB-D camera and add them to the simulation according to their ID, positions and dimensions wrt the camera frame with a sampling frequency of 15 Hz. These fruits are considered as harvesting targets for the robot. 

\begin{figure}[t]
	\centering {\centering\includegraphics[width=0.9\columnwidth]{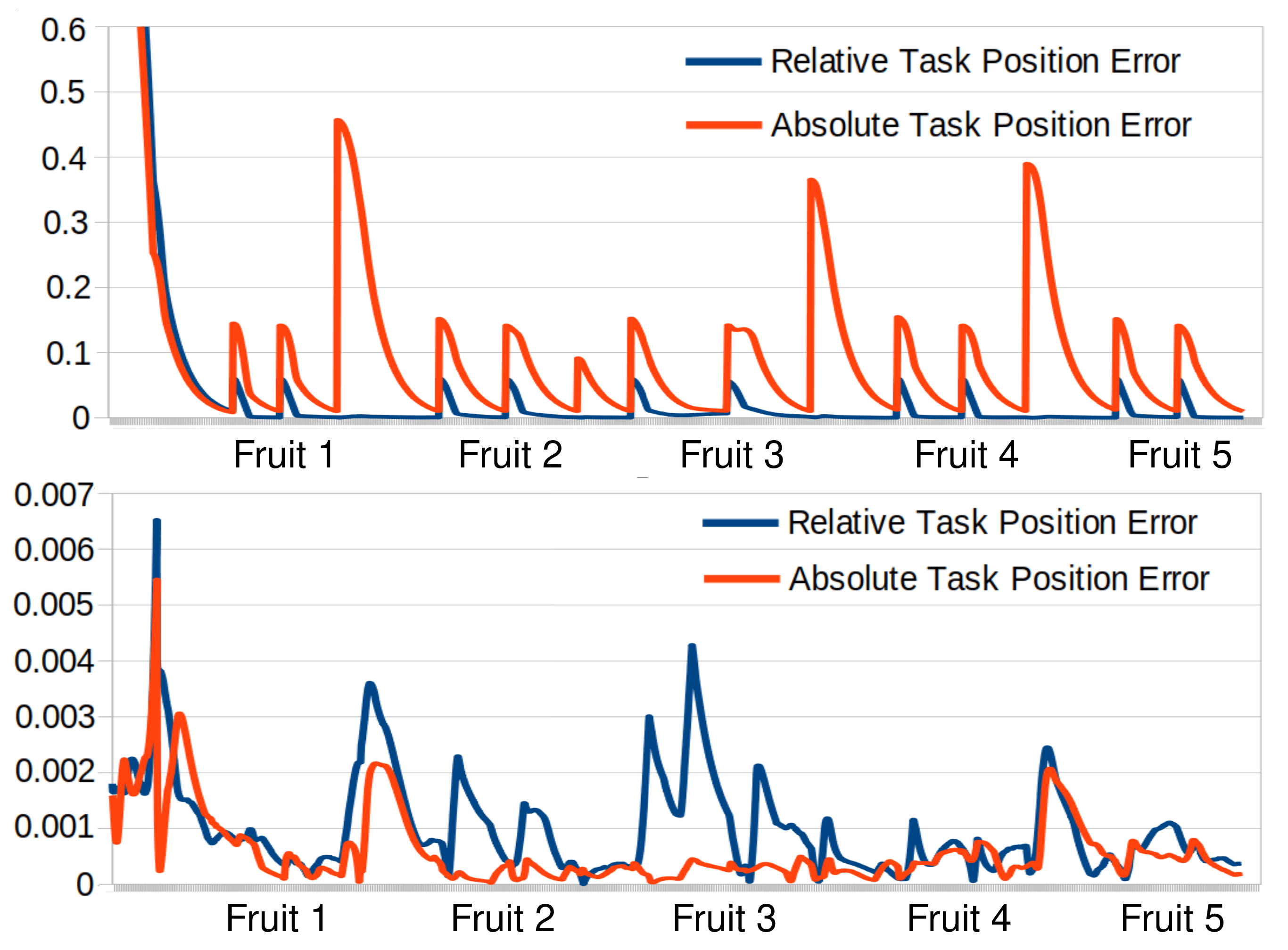}}
	\vspace{-0.1cm}
	\caption{Results of the harvesting experiment. Top: norm of the position error in meters. Bottom: norm of the orientation error in radian.}
	\vspace{-0.6cm}

	\label{error}
\end{figure}

In the first experiment, the camera detects 5 oranges in the scene as shown in Figure \ref{tree}. There is 1 undetected orange which is heavily occluded by leaves. We intentionally create obstacles and move fruits with our hands to test the performance of the fruit tracking. We validate our detection and tracking method on 1935 images. The fruit ID assignment is correct in 83\% of the 266 occluded images and it is 100\% accurate on the rest. Picking and collecting tools are simulated by a black cylinder and a yellow disk in CoppeliaSim. We assume that if both of the end-effectors reach their goal pose for the target fruit, the harvest is successful. The norms of the position error of the absolute and relative tasks in Figure \ref{error} show that the robot successfully harvests all the detected fruits in the scene. The experiment is finished in 32 seconds.

In the second experiment, the camera detects all 5 oranges in the scene and track them accurately at all times. The robot failed to harvest one orange due to its unreachable position. The rest of the oranges are successfully harvested.

In both experiments, we do not observe any collision between the robot and the tree trunk or oranges.

\section{CONCLUSION}\label{conclusion}
This paper describes a dual-arm autonomous harvesting system that uses state-of-the-art detection, tracking and control methods, which validated in an orange case study. Fruit detection shows promising results since all fruits except a heavily occluded one are detected correctly. Tracking results are positive despite the occlusions in the scene. Detected fruits in real world are successfully harvested in simulation. However, further experiments with the real robot and different fruits are necessary for the validation of our methods. During our real-robot experiments, we have encountered occlusions caused by robot arms which we could not test in our simulation. Also, since we do not consider branches as obstacles, interactions with them cause fruit displacements. 

In future, a more dynamic scene is needed to test the robustness of the tracking algorithm. In some real-world cases, fruits are not always facing downwards. To face this issue, a more precise detection strategy for the peduncle can be explored. A planning algorithm should efficiently determine harvesting path which can also reduce harvest time. 
Cutting and collecting strategies with tools are to be determined after real-robot experiments due to the difficulties on modeling physical interactions in robot simulators. 


The methods presented in this paper provide a step towards the goal of fully autonomous fruit harvesting systems. In this context, one arm solutions have already shown promising results. We believe that using dual-arm robots will improve harvesting performance and efficiency.

\bibliographystyle{IEEEtran}
\bibliography{ref}

\end{document}